\documentclass[sigconf]{acmart}

\usepackage{xcolor,colortbl}
\usepackage{multicol}
\usepackage{multirow}
\usepackage{caption}
\usepackage{subcaption}
\usepackage{graphicx}
\usepackage{hyperref}
\renewcommand\footnotetextcopyrightpermission[1]{} 
\acmConference[arxiv]{}{}{}

\begin{document}


\title[Multi-Prompt with Depth-Partitioned Cross-Modal Learning]{%
  Multi-Prompt with Depth-Partitioned Cross-Modal Learning
}


\author{Yingjie Tian}
\authornotemark[1]
\email{tyj@ucas.ac.cn}
\affiliation{%
  \institution{Research Center on Fictitious Economy and Data Science, University of Chinese Academy of Sciences}
  \country{China}
}

\author{Yiqi Wang}
\email{wangyiqi18@mails.ucas.edu.cn}
\affiliation{%
  \institution{School of Computer Science and Technology, University of Chinese Academy of Sciences}
  \city{Beijing}
  \country{China}
}
\author{Xianda Guo}
\email{xianda_guo@163.com}
\affiliation{%
   \institution{ Waytous}
  \country{China}
}

\author{Zheng Zhu}
\email{zhengzhu@ieee.org}
\affiliation{%
  \institution{PhiGent Robotics}
  \country{China}
}



\begin{abstract}
 In recent years, soft prompt learning methods have been proposed to fine-tune large-scale vision-language pretrained (VLP) models for various downstream tasks. These methods typically combine learnable prefix tokens with class tokens as inputs for models with frozen parameters. However, we argue that these methods often employ uni-prompts to describe the diverse prefixes of classes, which are suboptimal in that they fail to prompt classes' diverse attributes or other useful environmental information. This study introduces a partitioned multimodal multi-prompt (PMPO), a technique that extends a soft prompt from a single prompt to multiple prompts. To avoid the trivial solution, we creatively propose to divide the depths of a visual encoder and connect learnable prompts to the partitioned visual depths, enabling different prompts to capture the hierarchical contextual information of visual representations. Furthermore, to maximize the advantages of multi-prompt learning, we incorporate prior information derived from manually designed templates into the multi-prompt, thus improving the generalization ability of our approach. We evaluate the effectiveness of our approach on three standard tasks: new class generalization, cross-dataset evaluation, and domain generalization. For instance, our method achieves a $79.28$ harmonic mean averaged over 11 diverse image recognition datasets ($+7.62$ compared to that of 
CoOp
), demonstrating significant competitiveness compared to the state-of-the-art prompting methods. The code is available at \href{https://github.com/Wangyiqi/PMPO}{\textcolor{red}{PMPO}}.
\end{abstract}

\maketitle
\section{Introduction}
In recent years, large-scale vision-language pretrained (VLP) models have exhibited significant potential for various downstream tasks and datasets \cite{gao2021clip,zhou2022learning,zhou2022conditional,zhang2022pointclip}. Utilizing contrastive learning, these models leverage many text-image pairs to train text and image encoders \cite{radford2021learning}; e.g., 400 M text-image pairs from a website were used to train the contrastive language-image pretraining (CLIP) model. After completing pretraining, the text encoder transforms language prompts (e.g., "a photo of [class]") into text embeddings and calculates their similarities with visual embeddings for image class prediction.

Despite the benefits of language-guided supervision for exploring open-set visual concepts, these models often exhibit
sensitivity to manually designed template prompts for different datasets \cite{radford2021learning}. Recent studies have introduced various soft prompt learning methods \cite{zhou2022learning,zhou2022conditional}, which convert contextual prompt words into learnable vectors, demonstrating improved effectiveness and stability on downstream tasks compared to manually designed prompts.
\begin{figure}[htb]
\begin{center}
\includegraphics[width=1.0\linewidth]{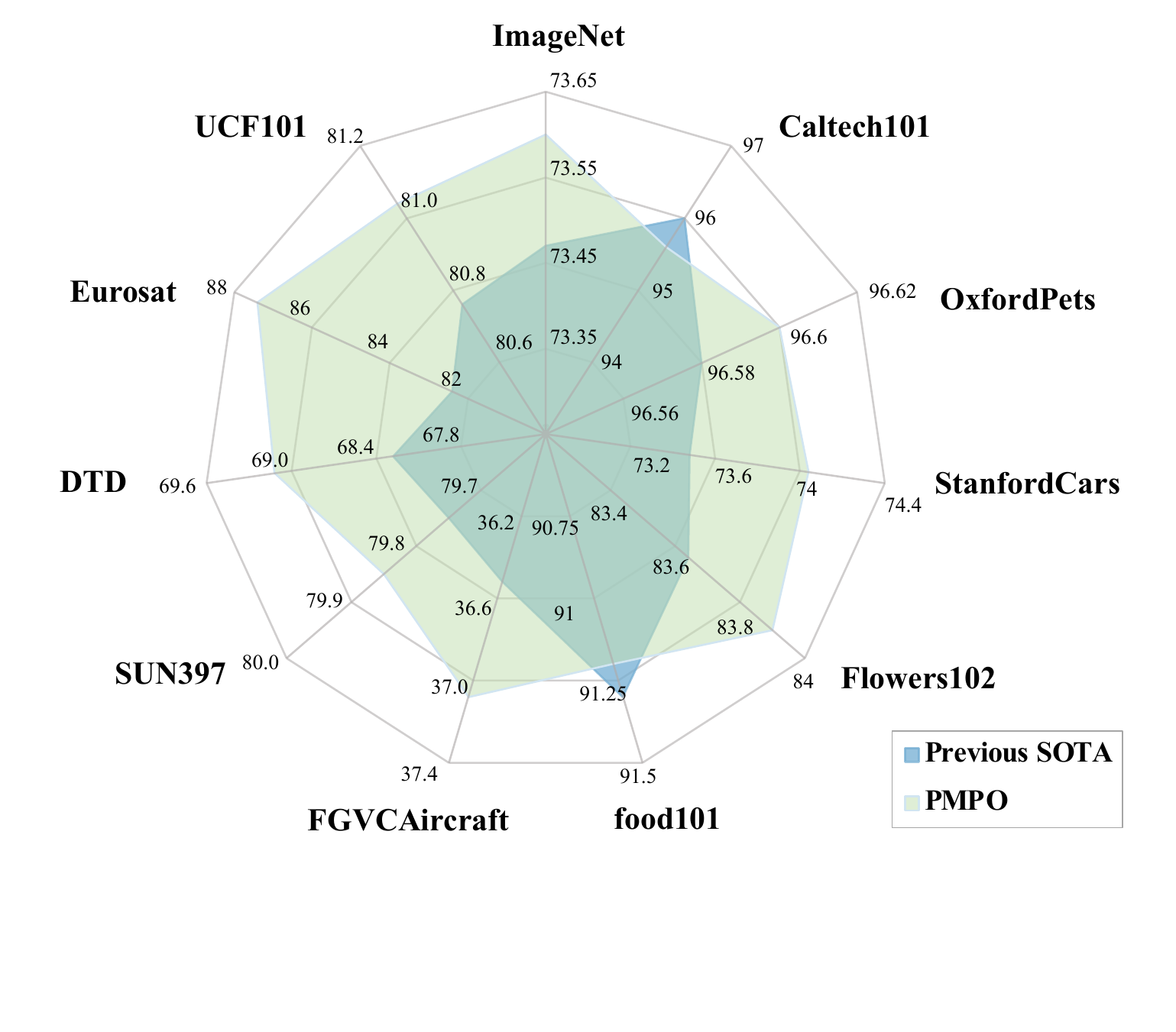}
\end{center}
\vspace{-1.2cm}
\caption{PMPO outperforms previous state-of-the-art models on 9 out of 11 diverse recognition datasets for the new classes generalization task.}
\label{fig:radar}
\end{figure}
\begin{figure*}[htbp]
\begin{center}
\includegraphics[width=1.00\linewidth]{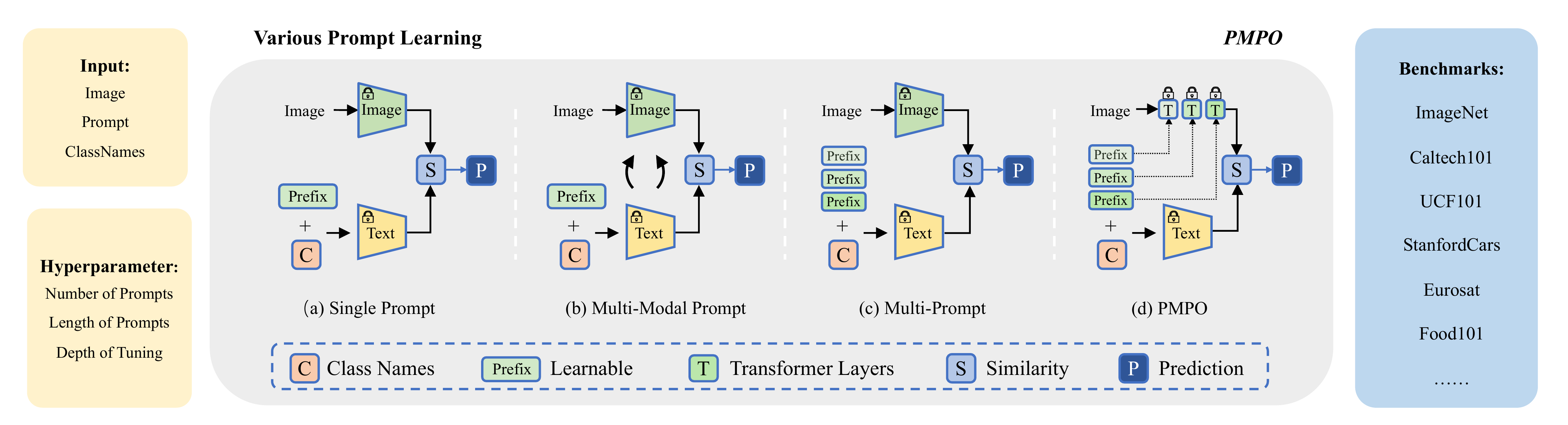}
\end{center}
\caption{Graphical illustration of different prompt learning methods. The image encoder and text encoder are pre-trained on large-scale image-text pairs and remain frozen during prompt tuning. This figure provides a brief comparison of single-prompt, multi-prompt, multi-modal prompt, and our proposed PMPO method.}
\label{fig:compare_prompt_learning}
\end{figure*}

\begin{figure*}[htbp]
\begin{center}
\includegraphics[width=1.0\linewidth]{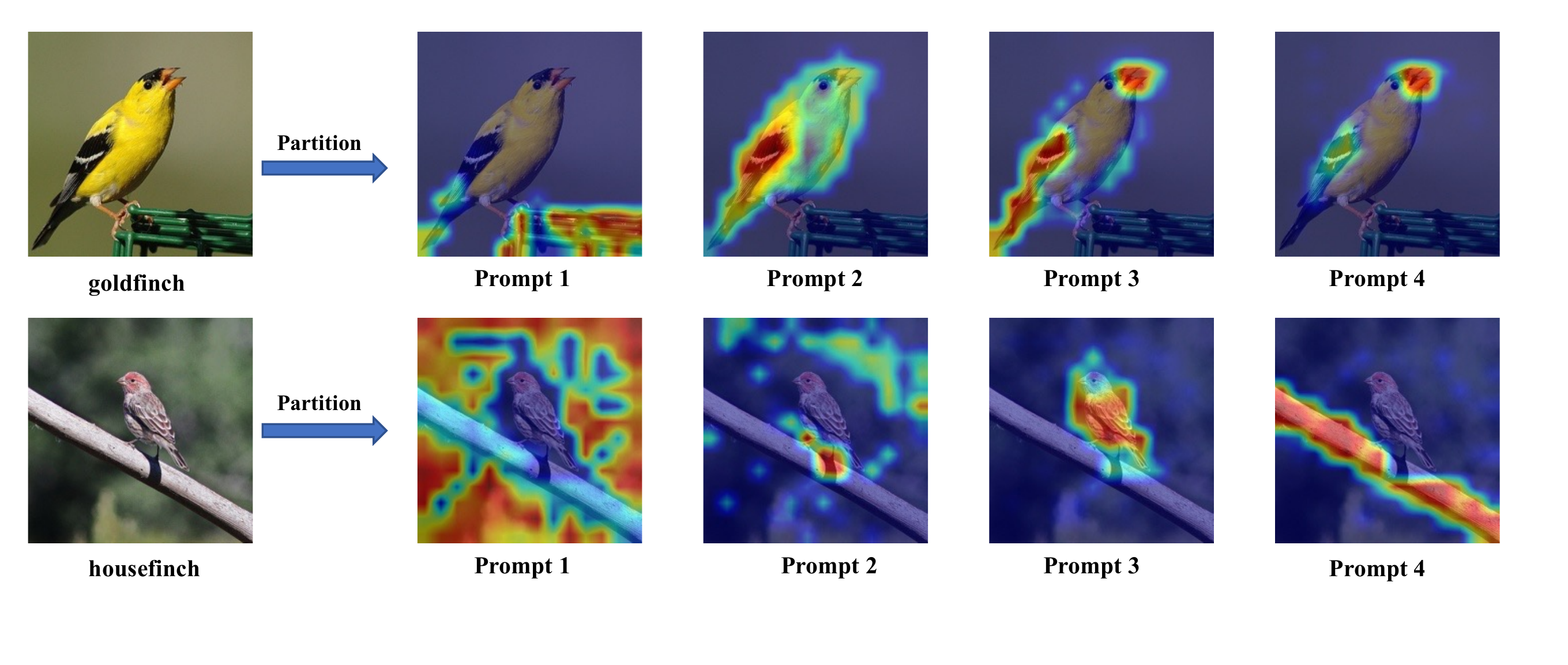}
\end{center}
\vspace{-0.9cm}
\caption{Key motivation of PMPO. We visualize the Score-CAM results for each prompt on two ImageNet samples. It is clear that different prompts indeed focus on distinct attributes of the classes. This observation highlights the importance of capturing diverse characteristics for effective recognition, as it allows the model to better understand various aspects of each class, leading to more accurate and robust predictions. }
\label{fig:score_cam}
\end{figure*}

Recently, some works have proven that ensembling multiple prefix templates can improve the accuracy of image classification. For example, the combination of two prefix templates, "[class] is a black and white striped animal" and "[class] looks like a horse," can effectively classify "zebra," "donkey," and "panda" better than using only one of them. However, less work has explored learning with multiple soft prompts. This work addresses the limitations of the existing soft prompt learning techniques. Standard prefix-based prompt methods learn a uni-prompt prefix, which is insufficient for extracting images' diverse characteristics. Therefore, we expand the uni-prompt to a multi-prompt to obtain more comprehensive implicit vision representations.  To learn a discriminative multi-prompt, a naive approach entails matching visual embeddings with the corresponding mean of multi-prompt class embeddings (Figure \ref{fig:compare_prompt_learning}:c). However, this naive method yields trivial solutions due to the absence of explicit differentiated training strategies for different prompts because each prompt is merely aligned with global visual embeddings. Therefore, the multi-prompt learns the same information rather than different information from the given images. To overcome this issue, we introduce a partitioned multimodal multi-prompt (PMPO), a multimodal technique that partitions the depths of a visual encoder and connects learnable prompts to the partitioned visual depths. This enables different prompts to learn the hierarchical contextual information of the observed visual representations. In Figure \ref{fig:score_cam}, we visualize Score-CAM results \cite{wang2020score} to elucidate the key motivation of the PMPO. The figure clearly shows that the PMPO's different prompts can focus on different information in the images.

We evaluate our method on 11 datasets covering diverse visual recognition tasks, including base class-to-new class generalization, cross-dataset evaluation, and domain generalization settings. The experimental results in Figure \ref{fig:radar} demonstrate that the PMPO surpasses the existing state-of-the-art prompt learning approaches, achieving absolute average gains in terms of both the base and new categories and the harmonic mean. Furthermore, the PMPO exhibits superior generalization capabilities and robustness under transfer and domain generalization settings. In summary, the primary contributions of this work are as follows.

We propose a multi-prompt with depth-partitioned cross-modal learning to leverage large-scale VLP models and extract the diverse characteristics of each input image.

To avoid trivial solutions, such as learning the same prompts in multi-prompt learning, we introduce depth-partitioned multimodal technology, which bridges the learnable prompts and visual contextual features at different depths.

We conduct extensive experiments on our proposed method, the PMPO, showcasing its exceptional performance across all few-shot recognition settings. These experiments highlight the comprehensive capabilities of the PMPO.
\section{Related Work}
\textbf{VLP Models: }
VLP models aim to establish cross-modal connections between the visual and language modalities through comprehensive pretraining \cite{chen2022pali,2022BLIP,li2023blip2}. Various pretraining objectives have been introduced, which allow the categorization of VLP methods into different groups based on their goals, including (masked) language modeling \cite{li2021align,wang2022image}, image-text contrastive learning \cite{radford2021learning,jia2021scaling,jain2021mural}, image-text matching \cite{li2021align,bao2022vlmo}, and combined approaches that merge the previously mentioned techniques \cite{2022BLIP,li2023blip2}. Most VLP approaches employ pretraining strategies using large-scale datasets containing image-text pairs. For example, the CLIP model \cite{radford2021learning} leverages 400 million image-text pairs in its contrastive learning process. Consequently, VLP models have demonstrated significant advancements in few-shot visual recognition tasks \cite{radford2021learning,gao2021clip,zhou2022learning,zhou2022conditional,zhang2022pointclip}, highlighting their capacity to provide improved open-world visual comprehension through language support.

\textbf{Prompt Learning:} Prompt learning was initially explored within the field of natural language processing (NLP) \cite{petroni2019language}. Recent studies have found that the performance of language models can be sensitive to input prompts \cite{lu2022fantastically,nie2022improving}, leading to the development of methods for improving prompt learning by using automatic prompt searches \cite{rubin2022learning,lu2022fantastically,shin-etal-2020-autoprompt} and prompt tuning \cite{lester-etal-2021-power}. Instead of manually creating prompts, Jiang et al. \cite{jiang2020can} used text mining and paraphrasing to generate a group of possible prompts, choosing the best prompts based on their training accuracy. Shin et al. \cite{shin2020autoprompt} presented AutoPrompt, an approach that uses gradients to find the most influential tokens in a vocabulary. Concurrently, recent research concerning contrastive learning-based vision-language models (VLMs) \cite{radford2021learning} has suggested learning virtual text as part of the texts utilized during the inference process.

Recently, CoOp \cite{zhou2022learning} and its related version CoCoOp \cite{zhou2022conditional} have integrated prompt learning into open-world visual understanding, enabling knowledge transfer from large-scale VLP models and significantly improving the performance of few-shot visual recognition tasks. MapLe \cite{khattak2022maple} employs learning across multiple transformer blocks in both the vision and language areas to progressively understand the combined behavior of both modalities. However, MapLe is a uni-prompt method and cannot provide an insightful explanation of its design. KgCoOp \cite{yao2023visual} boosts the generalization capacity of learnable prompts for unseen classes, reducing the gap between learnable and handcrafted prompts. The PLOT method \cite{chenplot} is a multi-prompt learning technique that enhances prompt learning by using the optimal transport distance \cite{monge1781memoire} to learn multiple local prompts, achieving more detailed vision-language matching. However, both of the above methods are uni-modal (only the language encoder of CLIP is used) approaches that lack cross-modal connections, which are crucial for effectively integrating visual and language information. A comparison between these techniques and our method is shown in Figure \ref{fig:compare_prompt_learning}. 
\begin{figure*}[htbp]
\begin{center}
 \includegraphics[width=1.0\linewidth]{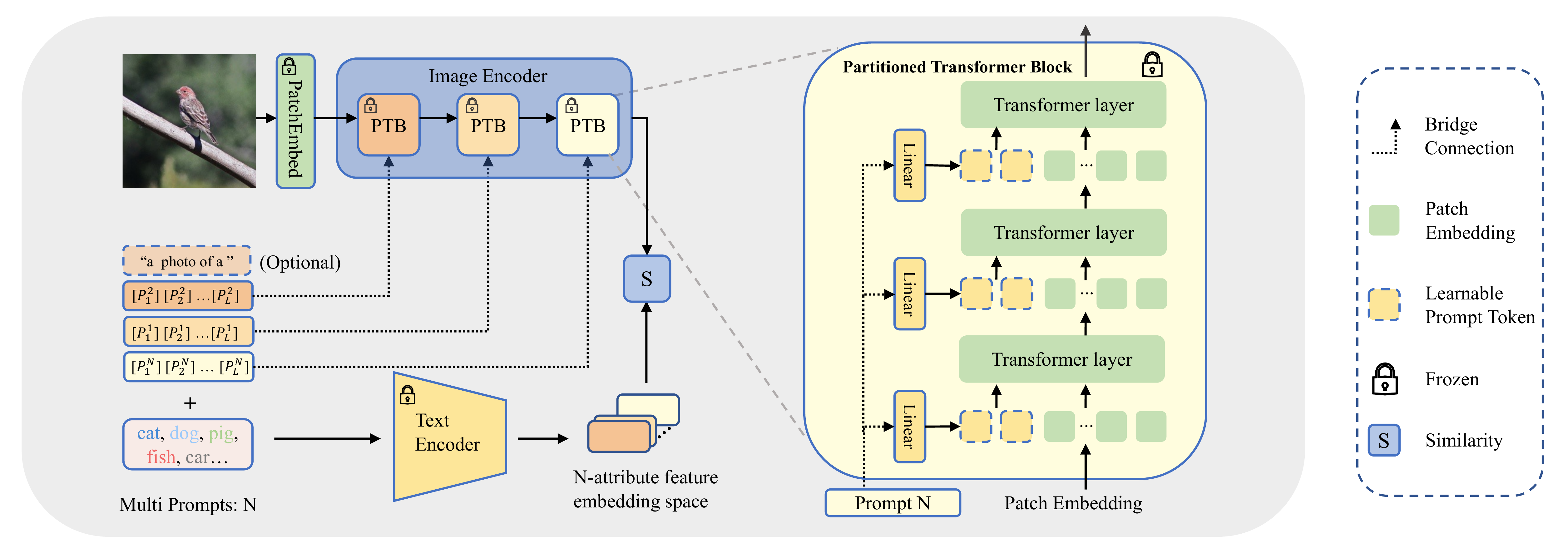}
\end{center}
\caption{Overview of the PMPO training pipeline: PMPO is based on the frozen CLIP. $N$ learnable prefix prompts establish bridge connections with Partitioned Transformer Blocks (PTB) enabling multi-prompts to learn hierarchical contextual visual representations. Within the PTB, Deep Visual Prompt Tuning is utilized for extracting visual embeddings. Furthermore, the ensemble of multi-prompts is computed within the feature embedding space, enhancing the overall effectiveness of the model.}
\label{fig:pipeline}
\end{figure*}

\section{Methods}
Figure \ref{fig:pipeline} shows an overview of our method. The PMPO first encodes multiple learnable language prompts using a pretrained text encoder (with an optional manually designed prompt). Then, the PMPO extracts visual features using a partitioned visual encoder. By default, the PMPO employs ViT-B \cite{dosovitskiy2020image} as its visual encoder and partitions its transformer blocks. For each partitioned transformer block, the PMPO utilizes the deep visual prompt tuning (D-VPT) \cite{jia2022visual} approach with visual context prompts projected from the corresponding learnable language prompt via linear layers. The final prediction is calculated by cosine similarity using the mean of the multi-prompt embeddings and visual embeddings. To better introduce the proposed method, we first briefly review the CLIP model \cite{radford2021learning}, which has demonstrated remarkable performance in various tasks. We then introduce the concepts of multi-prompt learning and partitioned multimodal prompt learning to extend the applicability of CLIP-like models.

\subsection{Preliminaries}
\textbf{Revisiting CLIP:} CLIP consists of two primary components: an image encoder (either a ResNet \cite{he2016deep} or a ViT \cite{dosovitskiy2020image}) and a text encoder (based on the transformer architecture). The image encoder's purpose is to transform the input images into feature embeddings, while the text encoder generates vectorized representations for word token sequences. During training, CLIP utilizes a contrastive loss function to learn a joint embedding space between images and text.

 Given a minibatch containing image-text pairs, CLIP aims to maximize the cosine similarity between each image and its corresponding text while minimizing the cosine similarities between the image and all other nonmatching texts. This process is symmetrically applied to the text as well. Once trained, CLIP can be employed to complete zero-shot image recognition tasks. For the image encoder, let $x$ represent the image features. Regarding the text encoder, let $\{w_i\}^{K}_{i=1}$ denote a set of produced weight vectors, each of which represents a category. Specifically, each $w_i$ originates from a prompt, such as "a photo of a \{class\}", where the $i$-th class name replaces the "\{class\}" token. The probability distribution over the class labels can be expressed as follows:
\begin{equation}
P(y|x) = \frac{\exp\left(sim(x,w_i)/{\tau}\right)}{\sum_{i=1}^K \exp\left(sim(x,w_i)/{\tau}\right)}
\end{equation}
In this equation, $sim()$ indicates cosine similarity, and $\tau$ refers to a learned temperature parameter.

\textbf{Prompt Learning:}
To better adapt to downstream tasks, CoOp \cite{zhou2022learning} was introduced as a novel approach that replaces the previous "a photo of a" template with $M$ learnable context vectors denoted as $p=[p_1, p_2, ..., p_M]$, which have the same dimensions as the corresponding word embeddings. For the $i$-th class, which is represented as $c_i$, the prompt changes to $t_i = \{p, c_i\}$. Here, $c_i$ represents the word embedding(s) for the relevant class name, and $p$ represents the context vectors shared across all classes (Figure \ref{fig:compare_prompt_learning}: a). The text encoder, denoted as $T(\cdot)$, is used to predict the probability of an input image $x$ belonging to a specific class $y$. The prediction probability is given by

\begin{equation}
P(y|x) = \frac{\exp\left(sim(x,T(t_i))/{\tau}\right)}{\sum_{i=1}^K \exp\left(sim(x,T(t_i))/{\tau}\right)}
\end{equation}

 CoOp employs a cross-entropy loss function for image recognition tasks and updates its context vectors during training. The base CLIP model remains frozen throughout this process.
\subsection{PMPO}
\textbf{Multi-Prompt Learning:} Although traditional soft prompt learning has achieved impressive performance in classification tasks, it relies on a uni-prompt to characterize the contexts of classes. To effectively capitalize on the full extent of the CLIP-pretrained knowledge, we introduce a set of N text prompts, denoted as $P^{'} = \{p^{n} | n = 1, \dots, N\}$. Each prompt, $P^{n} = [p_{1}^{n}, p_{2}^{n}, \dots, p_{M}^{n}]$, comprises exclusively trainable context tokens derived from the text encoder module, with $M$ representing the length of these tokens, which extends the uni-prompt from $(M, C)$ to $(N, M, C)$, as shown in Figure \ref{fig:compare_prompt_learning}: c.

The multi-prompts, denoted by $t^{'} = \{t^{n} | n = 1, \dots, N\}$ and $t^{n}_i = \{p^{n}, c_i\}$, are generated by prepending the text prompts $P^{'}$ to the $i$-th tokenized class names. Notably, all class names consistently share text prompts $P^{n}$. Here, $c_i$ corresponds to the word embedding associated with each class name.

We compute the mean of the text features in the embedding space to process these multi-prompts. Subsequently, the prediction probability is denoted by the following equation:
\begin{equation}
P(y|x) = \frac{\exp\left(sim(x,T^{\ast}_{i}/{\tau}\right)}{\sum_{i=1}^K \exp\left(sim(x,T^{\ast}_{i}/{\tau}\right)}
\label{multi-prediction}
\end{equation}
\begin{equation}
    T^{\ast}_{i}=\mathrm{\underset{n=1\dots N}{mean}}(T(t^{n}_i))
\end{equation}
Utilizing multi-prompts offers advantages over the uni-prompt approach, particularly regarding their ability to capture diverse visual representations of an image. This provides more comprehensive representations and yields improved model performance.

\textbf{Depth-Partitioned Cross-Modal Learning:}
Directly using multi-prompts can lead to trivial solutions, as there is no explicit guide to maintain the diversity of multi-prompts. To address this issue, we propose depth-partitioned cross-modal learning. We use a linear projection $f(\cdot)$ for each prompt to map it to different levels of the image encoder blocks to make different prompts adapt to the different levels of visual representations produced for an image. This is an effective way to maintain prompt diversity.

In the case of CLIP, we assign the $N$ learnable prompts to various transformer blocks. We set the depth of the transformer to $D$, and each text prompt is assigned to $D/N$ transformer layers. We compute a sequence of vectors $\{V_1, V_2, \dots, V_D\}$ using linear layers $g$ with text prompts $\{p^1, p^2, \dots, p^N\}$ as inputs: \begin{equation}
V_{1},V_{2}, \dots ,V_{D}=g(p^{1}, \dots ,p^{N})
\end{equation} We then use these vectors to compute a series of embeddings $[x_i, E_i]$ using a set of transformer layers $L_i$, : \begin{equation}
[x_i, E_i] = L_i([x_{i-1}, V_{i-1}, E_{i-1}]), \quad i = 1, 2, \dots, D
\end{equation} Here, $x_i$ is the embedding for the $[CLS]$ and $E_{i}$ denotes the visual embedding tokens. $[\cdot,\cdot]$ indicates stacking and concatenation, and the resulting embedding $[x_i, V_i, E_i]$ has dimensions of $R^{(1+M+m)\times d}$. Each layer $L_i$ in the network comprises multiheaded self-attention (MSA), feed-forward networks (FFNs), layer normalization, and residual connections. We map the final layer's $[CLS]$ embedding $x$ to a predicted class probability distribution using Equation \ref{multi-prediction}, which can be replaced by the conditioned $x(p^1,p^2,\dots,p^N)$:
\begin{equation}
P(y|x) = \frac{\exp\left(sim(x(p^1,p^2 \dots p^N),T^{\ast}_{i}/{\tau}\right)}{\sum_{i=1}^K \exp\left(sim(x(p^1,p^2 \dots p^N),T^{\ast}_{i}/{\tau}\right)}
\end{equation} $T^{\ast}_{i}$ is the mean of the encoded text embeddings.
\begin{table*}[htbp]
    \centering
    \caption{Comparison of our approach with state-of-the-art methods in the base-to-new generalization setting,utilizing $N=4$ learnable prompts with a manual prompt. All results are trained on 16-shot samples from the base classes, with the Vit-B/16 backbone architecture. H:Harimonic mean}
    \setlength{\tabcolsep}{4pt}
    \begin{subfigure}[b]{0.23\textwidth}
        \centering
        \caption{Average over 11 datasets}
        \begin{tabular}{p{1cm}p{0.5cm}p{0.7cm}|c}
            \toprule
             & Base & New & H \\
            \midrule
            CLIP & 69.34 & 74.22 & 71.70 \\
            CoOp & 82.69 & 63.22 & 71.66 \\
            CoCoOp & 80.47 & 71.69 & 75.83 \\
            MaPLe & 82.27 & 75.14 & 78.55 \\
            KgCoOp & 80.73 & 73.60 & 77.00 \\
            \midrule
            PMPO & \textbf{82.91} & \textbf{75.95}& \textbf{79.27}\\
            \bottomrule
        \end{tabular}
        
    \end{subfigure}
    \hspace{1cm}
    \begin{subfigure}[b]{0.23\textwidth}
        \centering
        \caption{ImageNet}
        \begin{tabular}{p{1cm}p{0.5cm}p{0.7cm}|c}
            \toprule
             & Base & New & H \\
            \midrule
            CLIP & 72.43 & 68.14 & 70.22 \\
            CoOp & 76.47 & 67.88 & 71.92 \\
            CoCoOp & 75.98 & 70.43 & 73.10 \\
            MaPLe & 76.66 & 70.54 & 73.47 \\
            KgCoOp & 75.83 & 69.96 & 72.78 \\
            \midrule
            PMPO & \textbf{76.94} & \textbf{70.55}& \textbf{73.60}\\
            \bottomrule
        \end{tabular}
    \end{subfigure}
\hspace{1cm}
\begin{subfigure}[b]{0.23\textwidth}
        \centering
        \caption{Caltech101}
        \begin{tabular}{p{1cm}p{0.5cm}p{0.7cm}|c}
            \toprule
             & Base & New & H \\
            \midrule
            CLIP & 96.84 & 94.00 & 95.40 \\
            CoOp & 98.00 & 89.81 & 93.73 \\
            CoCoOp & 97.96 & 93.81 & 95.84 \\
            MaPLe & 97.74 & 94.36 & 96.02 \\
            KgCoOp & 97.72 & \textbf{94.39} & \textbf{96.03} \\
            \midrule
            PMPO & \textbf{98.04} & 93.34& 95.63\\
            \bottomrule
        \end{tabular}
\end{subfigure}
\hspace{1cm}
\begin{subfigure}[b]{0.23\textwidth}
        \centering
        \caption{OxfordPets}
        \begin{tabular}{p{1cm}p{0.5cm}p{0.7cm}|c}
            \toprule
             & Base & New & H \\
            \midrule
            CLIP & 91.17 & 97.26 & 94.12 \\
            CoOp & 94.24 & 96.66 & 95.43 \\
            CoCoOp & 95.20 & 97.69 & 96.43 \\
            MaPLe & 95.43 & 97.76 & 96.58 \\
            KgCoOp & 94.65 & \textbf{97.76} & 96.18 \\
            \midrule
            PMPO & \textbf{95.84} & 97.37& \textbf{96.60}\\
            \bottomrule
        \end{tabular}
        
\end{subfigure}
\hspace{1cm}
\begin{subfigure}[b]{0.23\textwidth}
        \centering
        \caption{StanfordCars}
        \begin{tabular}{p{1cm}p{0.5cm}p{0.7cm}|c}
            \toprule
             & Base & New & H \\
            \midrule
            CLIP & 63.37 & 74.89 & 68.65 \\
            CoOp & \textbf{78.12} & 60.40 & 68.13 \\
             CoCoOp & 70.49 & 73.59 & 72.01 \\
            MaPLe & 72.94 & 74.00 & 73.47 \\
            KgCoOp & 71.76 & \textbf{75.04} & 73.36 \\
            \midrule
            PMPO & 74.16 & 74.15& \textbf{74.16}\\
            \bottomrule
        \end{tabular}

\end{subfigure}
\hspace{1cm}
\begin{subfigure}[b]{0.23\textwidth}
        \centering
        \caption{Flowers102}
        \begin{tabular}{p{1cm}p{0.5cm}p{0.7cm}|c}
            \toprule
             & Base & New & H \\
            \midrule
            CLIP & 72.08 & \textbf{77.80} & 74.83 \\
            CoOp & \textbf{97.60} & 59.67  & 74.06 \\
            CoCoOp & 94.87 & 71.75 & 81.71 \\
            MaPLe & 95.92 & 72.46 & 82.56 \\
            KgCoOp & 95.00 & 74.73  & 83.65 \\
            \midrule
            PMPO & 97.12 & 73.85& \textbf{83.89}\\
            \bottomrule
        \end{tabular}
        
\end{subfigure}
\hspace{1cm}
\begin{subfigure}[b]{0.23\textwidth}
        \centering
         \caption{Food101}
        \begin{tabular}{p{1cm}p{0.5cm}p{0.7cm}|c}
            \toprule
             & Base & New & H \\
            \midrule
            CLIP & 90.10 & 91.22 & 90.67 \\
            CoOp & 88.33 & 82.26  & 85.19 \\
            CoCoOp & 90.70 & 91.29 & 90.99 \\
            MaPLe & \textbf{90.71} & \textbf{92.05} & \textbf{91.38} \\
            KgCoOp & 90.05 & 91.70  & 91.09 \\
            \midrule
            PMPO & 90.58 & 91.86 & 91.22\\
            \bottomrule
        \end{tabular}
       
\end{subfigure}
\hspace{1cm}
\begin{subfigure}[b]{0.23\textwidth}
        \centering
        \caption{FGVCAircraft}
        \begin{tabular}{p{1cm}p{0.5cm}p{0.7cm}|c}
            \toprule
             & Base & New & H \\
            \midrule
            CLIP & 27.19 & \textbf{36.29} & 31.09 \\
            CoOp & \textbf{40.44} & 22.30  & 28.75 \\
            CoCoOp & 33.41 & 23.71 & 27.74 \\
            MaPLe & 37.44 & 35.61 & 36.50 \\
            KgCoOp & 36.21& 33.55  & 34.83 \\
            \midrule
            PMPO & 38.46 & 35.99 & \textbf{37.18}\\
            \bottomrule
        \end{tabular}
        
\end{subfigure}
\hspace{1cm}
\begin{subfigure}[b]{0.23\textwidth}
        \centering
        \caption{SUN397}
        \begin{tabular}{p{1cm}p{0.5cm}p{0.7cm}|c}
            \toprule
             & Base & New & H \\
            \midrule
            CLIP & 69.36 & 75.35 & 72.23 \\
            CoOp & 80.60 & 65.89  & 72.51 \\
            CoCoOp & 79.74 & 76.86 & 78.27 \\
            MaPLe & 80.82 & \textbf{78.70} & 79.75 \\
            KgCoOp & 80.29 & 76.53  & 78.36 \\
            \midrule
            PMPO & \textbf{81.54} & 78.22 & \textbf{79.85}\\
            \bottomrule
        \end{tabular}
\end{subfigure}
\hspace{1cm}
\begin{subfigure}[b]{0.23\textwidth}
        \centering
         \caption{DTD}
        \begin{tabular}{p{1cm}p{0.5cm}p{0.7cm}|c}
            \toprule
             & Base & New & H \\
            \midrule
            CLIP & 53.24 & 59.90 & 56.37 \\
            CoOp & 79.44 & 41.18  & 54.24 \\
            CoCoOp & 77.01 & 56.00 & 64.16 \\
            MaPLe & \textbf{80.36} & 59.18 & 68.16 \\
            KgCoOp & 77.55 & 54.99  & 64.35 \\
            \midrule
            PMPO & 80.21 & \textbf{60.95} & \textbf{69.27}\\
            \bottomrule
        \end{tabular}
\end{subfigure} 
\hspace{1cm}
\begin{subfigure}[b]{0.23\textwidth}
        \centering
        \caption{Eurosat}
        \begin{tabular}{p{1cm}p{0.5cm}p{0.7cm}|c}
            \toprule
             & Base & New & H \\
            \midrule
            CLIP & 56.48 & 64.05 & 60.03 \\
            CoOp & 92.19 & 54.74  & 68.69 \\
            CoCoOp & 87.49 & 60.04 & 71.21 \\
            MaPLe & 94.07 & 73.23 & 82.35 \\
            KgCoOp & 85.64 & 64.34  & 73.48 \\
            \midrule
            PMPO & \textbf{94.24} & \textbf{81.85} & \textbf{87.41}\\
            \bottomrule
        \end{tabular}
\end{subfigure}
\hspace{1cm}
\begin{subfigure}[b]{0.23\textwidth}
        \centering
        \caption{UCF101}
        \begin{tabular}{p{1cm}p{0.5cm}p{0.7cm}|c}
            \toprule
             & Base & New & H \\
            \midrule
            CLIP & 7530. & 77.50 & 73.85 \\
            CoOp & 84.69 & 56.05  & 67.46 \\
            CoCoOp & 82.33 & 73.45 & 77.64 \\
            MaPLe & 83.00 & \textbf{78.66} & 80.77 \\
            KgCoOp & 82.89 & 76.67  & 79.65 \\
            \midrule
            PMPO & \textbf{84.85} & 77.64 & \textbf{81.09}\\
            \bottomrule
        \end{tabular}
\end{subfigure}
\label{tab:base_to_new}
\end{table*}

\textbf{Ensembling Manual Prompts:} Another advantage of multi-prompt learning lies in its ability to utilize manually created prompts as prior information, which can be combined with multiple learnable prompts to generate the final text features. As detailed in CoCoOp \cite{zhou2022conditional}, learnable prompts can yield improved classification performance for base classes but may often result in overfitting. Conversely, the vanilla CLIP method \cite{radford2021learning} typically demonstrates superior generalization capabilities when employing manual prompts. Thus, our framework can benefit from both manual and learnable prompts by employing the following formulation:
\begin{equation} T^{\ast}_{i}=mean([T(t^{1}_i),T(t^{2}_i),\dots,T(t^{N}_i),T(t^{prior}_{i})])
\end{equation}
\section{Experiments}


\textbf{Datasets:} In our study, we focus on two primary settings: base class-to-new class generalization and cross-dataset transfer. We utilize 11 image recognition datasets from Zhou et al.\cite{zhou2022learning}, encompassing a wide array of recognition tasks. The benchmarks include ImageNet \cite{deng2009imagenet} and Caltech101 \cite{fei2004learning} for generic object classification; OxfordPets \cite{parkhi2012cats}, StanfordCars \cite{krause20133d}, Flowers102 \cite{nilsback2008automated}, Food101 \cite{bossard2014food}, and FGVCAircraft \cite{maji2013fine} for fine-grained classification; SUN397\cite{xiao2010sun} for scene recognition; UCF101 \cite{soomro2012ucf101} for action recognition; DTD \cite{cimpoi2014describing} for texture classification; and EuroSAT \cite{helber2019eurosat} for satellite imagery recognition.

To conduct domain generalization experiments, we employ ImageNet as the source dataset and four ImageNet variants with distinct domain shifts as the target datasets: ImageNetV2 \cite{recht2019imagenet}, ImageNet-Sketch \cite{wang2019learning}, ImageNet-A \cite{hendrycks2021natural}, and ImageNet-R \cite{hendrycks2021many}. In line with CoOp, we randomly sample a 16-shot training set for each dataset and use the original test set for evaluation purposes. The results are averaged across three runs implemented with different seeds.

\textbf{Training Details:} In our experiments, we employ ViT-B/16 \cite{zhou2022learning} as the backbone of CLIP. We adopt a few-shot training strategy with 16 shots randomly sampled using different seeds. For the base class-to-new class generalization experiment, we set the number of prompts to 4 by default, the bridge depth to 12 (considering both vision transformer layers), and the contextual token length to 10. Regarding the other hyperparameters, we follow CoOp's configuration and use the stochastic gradient descent (SGD) optimizer with an initial learning rate of 0.002 and a batch size of 8 with 6 epochs for most datasets, except for ImageNet, which maintains a learning rate of 0.01 and a batch size of 32. The learning rate is decayed using cosine annealing, and the optimizer warms up with a 1e-5 learning rate.
For the domain transfer and cross-dataset experiments, we adopt CoCoOp's settings, which suggest that shorter context lengths yield better performance and stronger robustness to domain shift. Thus, we set the token length to 4, and the learning rate is set to 0.01 with 2 epochs. The other parameters remain the same as those used in the base class-to-new class experiments. All experiments are conducted on 4 Nvidia RTX3090 GPUs. For a fair comparison, all the results are averaged over three different seeds.
\subsection{Generalization From Base Classes to New Classes}
We build upon the work of CoOp\cite{zhou2022learning} and divide each of the 11 recognition datasets into two parts: base classes and new classes. We train the PMPO on the base classes and evaluate its performance on the new classes, comparing our method with five baselines: CLIP\cite{radford2021learning}, CoOp\cite{zhou2022learning}, CoCoOp\cite{zhou2022conditional}, MapLe\cite{khattak2022maple}, and KgCoOp\cite{yao2023visual}. Here, we provide a brief overview of these five baseline methods. CLIP\cite{radford2021learning} utilizes a manually designed template, typically employing "a photo of [cls]" as the text encoder prompt. CoOp\cite{zhou2022learning} replaces the manually designed prompt with a uni-learnable prompt. CoCoOp\cite{zhou2022conditional} incorporates a conditioned uni-learnable prompt. MapLe\cite{khattak2022maple} is a state-of-the-art approach that uses multimodal language-vision prompts. KgCoOp\cite{yao2023visual} leverages prior information from fixed, manually designed CLIP prompts to train learnable prompts via knowledge distillation. The results of our method and the baselines are presented in Table \ref{tab:base_to_new}.

\textbf{Significant Improvement for Base Classes:} As shown in Table \ref{tab:base_to_new}, our proposed PMPO achieves the best performance on the base classes for 6 out of 11 datasets and the best average performance across the base classes when compared to the baseline methods. Generally, achieving higher performance on base classes can result in overfitting and reduced generalizability to new classes. Among the existing state-of-the-art methods, CoOp is the best-performing previously developed method on the base classes, but it typically exhibits an overfitting problem, as argued by CoCoOp. However, the PMPO significantly outperforms CoOp in terms of their averages produced on 11 datasets ($82.69\%$ vs. $82.91\%$) while improving its ability to generalize to new classes. This exceptional performance demonstrates the PMPO's robust capability to adapt to downstream tasks based on a pretrained CLIP model. In the ablation study (Section \ref{section:ablation}), we will discuss the relationship between increasing the number of prompts and improving the accuracy achieved for the base classes.
\begin{figure}[tbp]
\begin{center}
\includegraphics[width=1.0\linewidth]{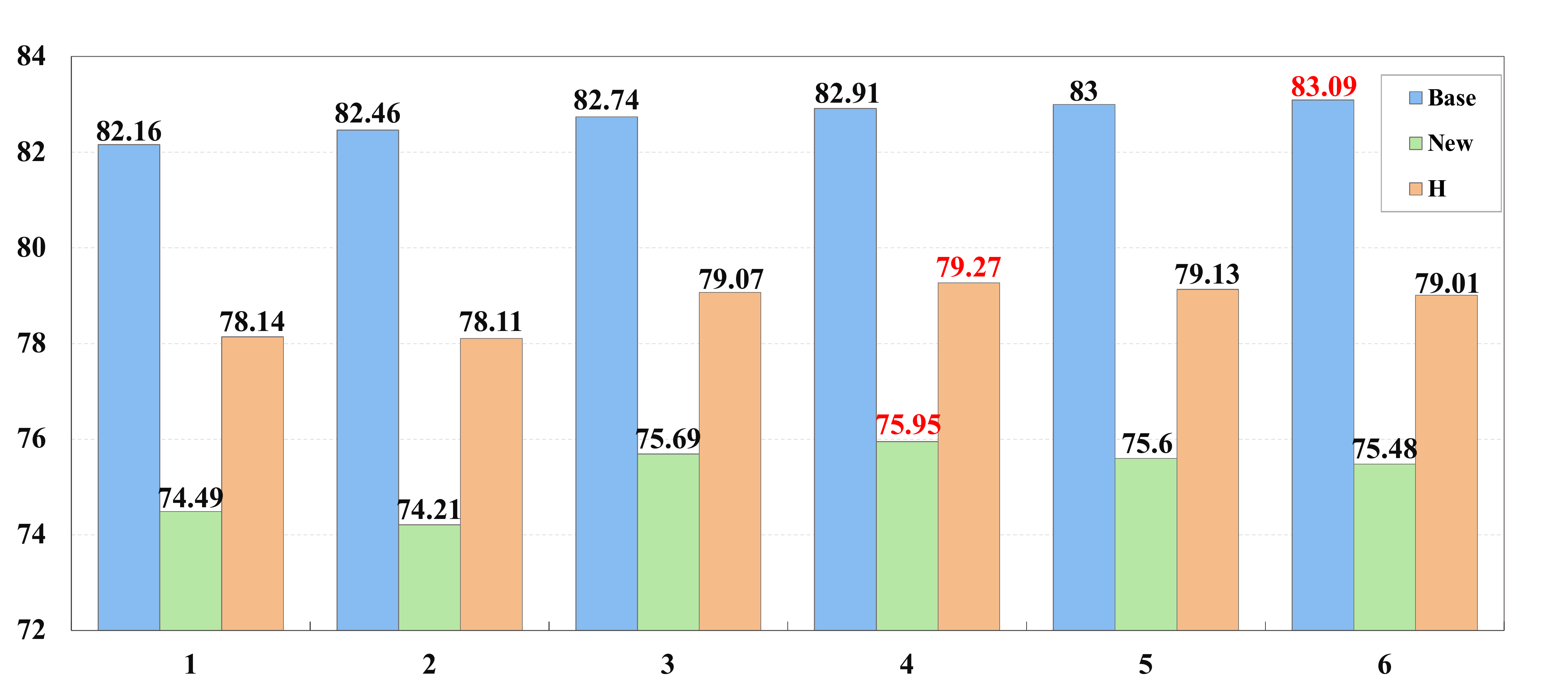}
\end{center}
\vspace{-0.3cm}
\caption{Ablation number of prompts}
\vspace{-0.5cm}
\label{fig:ablation_num}
\end{figure}

\textbf{Generalization to New Classes:} With improvements achieved for base classes, the PMPO also exhibits enhanced generalization to new classes in terms of the average unseen accuracy attained across all 11 datasets. Specifically, CLIP, KgCoOp, MapLe, and the PMPO achieve the best unseen accuracy rates on 2/11, 3/11, 3/11, and 3/11 of the datasets, respectively. While CoOp overfits the base classes, CoCoOp improves its generalization ability through conditional image context learning, while KgCoOp enhances its ability using knowledge-guided context optimization. However, they all underperform compared to CLIP on the new classes. In contrast, the PMPO surpasses CLIP on 7/11 datasets and demonstrates a $1.73\%$ average improvement across datasets for the new classes. The PMPO also performs better in terms of generalization than the previous best method, MapLe ($75.12\%$ vs. $75.95\%$), in the new class generalization task. Notably, the PMPO significantly outperforms MapLe on Eurosat \cite{helber2019eurosat}, with an impressive $8.52\%$ improvement. When considering the base and new classes, the PMPO achieves the best harmonic mean on 9 out of 11 datasets and attains the highest average performance, underlining its effectiveness and robustness in adapting to diverse tasks.
\begin{table*}[htbp]
  \centering
  \caption{ Comparison of PMPO with other methods in domain generalization setting.}
    \setlength{\tabcolsep}{15pt}
    \begin{tabular}{lccccc}
    \toprule
    \midrule
    & \multicolumn{1}{c}{Source} & \multicolumn{4}{c}{Target} \\
     \cmidrule{2-6}    
    & {Imagnet} & {ImageNetV2} & {ImageNet-Sketch} & {ImageNet-A} & {ImageNet-R}    \\
    \midrule  
    CLIP & 66.73 &60.83 & 46.15 & 47.77 & 73.96 \\
    CoOp & \textbf{71.51} & 64.20 & 47.99 & 49.71 & 75.21 \\
    CoCoOp & 71.02 & 64.07 & 48.75& 50.63 & 76.18 \\
    MaPLe & 70.72 & 64.07 & 49.15 & \textbf{50.90} & 76.98 \\
    KgCoOp & 71.20& 64.20 & 48.97& 50.69 & 76.70 \\
    \midrule   
    PMPO  & 70.78 & \textbf{64.20} & \textbf{49.64} & 50.63 & \textbf{77.12} \\
    \midrule
    \bottomrule
    \end{tabular}%
  \label{tab: domain}%
\end{table*}

\subsection{Ablation Study}
\textbf{Number of Prompts:}
To analyze the impact of the number of prompts, we conduct experiments on 11 datasets. We compare the average base, new, and harmonic mean accuracy performances of different methods, as shown in Figure \ref{fig:ablation_num}. All experiments are conducted with a depth $D=12$ and a length $L=10$ while varying the number of prompts from $N=1$ to $N=6$. A further analysis of the results reveals a continuous improvement in the base accuracy performance as the number of prompts increases (from $82.16\%$ to $83.09\%$), which demonstrates the enhanced fitting ability of the multi-prompt strategy. However, for new classes, the generalization performance increases when the number of prompts ranges from $N=1$ to $N=4$. When $N>4$, the PMPO begins to overfit the base classes, leading to a decline in the performance achieved for unseen classes. Considering both base and new classes, $N=4$ is clearly the optimal setting.
\begin{table}[htbp]
  \caption{Effect of ensembling the manually-designed prompt on the base-to-new  generalization on $11$ datasets. $M$ means  the manually-designed prompt. H:Harmonic mean.}
  \centering
    \setlength{\tabcolsep}{3.5pt}
    \begin{tabular}{lcc|cc|cc}
    \toprule
    \multicolumn{1}{l}{} & \multicolumn{2}{c|}{Base}  & \multicolumn{2}{c|}{New}&  \multicolumn{2}{c}{H}  \\
    \cmidrule{2-7}
    \multicolumn{1}{l}{datasets} & \multicolumn{1}{c}{w/o M}  & \multicolumn{1}{c|}{w/i M}& \multicolumn{1}{c}{w/o M}& \multicolumn{1}{c|}{w/i M} & \multicolumn{1}{c}{w/o M} & \multicolumn{1}{c}{w/i M}  \\
     \midrule   
     Imagnet & \textbf{77.11} &76.94 &70.45 &\textbf{70.55} & \textbf{73.62} &73.60   \\
     Caltech101 & \text{98.34} & 98.04& 93.27 &\textbf{93.24} & \textbf{95.74} & 95.63   \\
     OxfordPets & 95.78 &\textbf{95.84} &\textbf{97.45} & 97.37 & \textbf{96.61} & 96.60   \\
     StanfordCars & \textbf{74.76} &74.16 &74.06 &\textbf{74.15} &\textbf{74.41} & 74.16 \\
     Flowers102 & 96.68 & \textbf{97.12}& 71.87&\textbf{73.85} & 83.36 & \textbf{83.89} \\
     Food101 & 90.52 & \textbf{90.58}&91.50 & \textbf{91.80} & 91.00 & \textbf{91.22} \\
     FGVCAircraft & 38.14  & \textbf{38.46} &34.95 & \textbf{35.99} &36.47 &\textbf{37.18} \\
     SUN397 & 81.54 &\textbf{81.71} &77.80 &\textbf{78.22} & 79.71& \textbf{79.85} \\
     DTD & \textbf{81.33} &80.21 &59.54 &\textbf{60.95} &68.75 &\textbf{69.27}  \\
     EuroSAT & 93.73 &\textbf{94.24} &78.14 &\textbf{81.85} & 85.23&\textbf{87.41}   \\
     UCF101 &  \textbf{85.06}&84.85 & 76.74&\textbf{77.64} &80.69 &\textbf{81.09} \\
     \midrule 
     Average & \textbf{83.01}&75.09 &82.91 &\textbf{75.95} &78.85 &\textbf{79.27}\\
    \midrule         
    \bottomrule
    \end{tabular}%
  \label{tab: ablation_m2}
\end{table}
\begin{table}[htbp]
  \caption{Ablation on alternative prompt learning design. SS:Single-Prompt with Sing-Modal, MS: Multi-Prompt with Sing-Modal, SM: Single-Prompt with Multi-Modal, H: Harmonic mean.}
  \centering
    \setlength{\tabcolsep}{17pt}
    \begin{tabular}{lccc}
    \toprule
    \multicolumn{1}{l}{Method} & \multicolumn{1}{c}{Base}  & \multicolumn{1}{c}{New}&  \multicolumn{1}{c}{H}  \\
    \midrule 
     CoOp-SS & 82.69& 63.22&71.66 \\
     PMPO-MS  & 81.22 & 73.23&  77.27   \\
     PMPO-SM  &82.16 & 74.49& 78.14   \\
     \midrule 
     PMPO &  \textbf{82.91}&\textbf{75.95 }&\textbf{79.27}  \\
    \midrule         
    \bottomrule
    \end{tabular}%
  \label{tab: ablation_modal}
\end{table}
\begin{table*}[htbp]
  \caption{ Comparison of PMPO in the cross-dataset transfer setting with other state-of-the-art methods. Prompts are trained on ImageNet with 16-shot per classes and applied to the 10 target datasets. Clearly, CoCoOp demonstrates better transferability than other methods. }
  \centering
    \setlength{\tabcolsep}{7pt}
    \begin{tabular}{lcccccccccccc}
    \toprule
    \midrule 
    & \multicolumn{1}{c}{Source} & \multicolumn{11}{c}{Target} \\
     \cmidrule{2-13}    
    & \rotatebox[origin=c]{90}{Imagnet} & \rotatebox[origin=c]{90}{Caltech101} & \rotatebox[origin=c]{90}{OxfordPets} & \rotatebox[origin=c]{90}{StanfordCars} & \rotatebox[origin=c]{90}{Flowers102} & \rotatebox[origin=c]{90}{Food101} & \rotatebox[origin=c]{90}{Aircraft} & \rotatebox[origin=c]{90}{SUN397} & \rotatebox[origin=c]{90}{DTD} & \rotatebox[origin=c]{90}{EuroSAT} & \rotatebox[origin=c]{90}{UCF101} & \rotatebox[origin=c]{90}{Average}    \\
    \midrule   
    CoOp& \textbf{71.51}& 93.70 & 89.14 & 64.51&68.71&85.30&18.47 & 64.15 & 41.92 & 46.39 & 66.55&63.88 \\
    CoCoOp & 71.02 & \textbf{94.43} & 90.14 & 65.32 & 71.88 & 86.06 & 22.94 & 67.36 & 45.73 & 45.37 & 68.21 & 65.74 \\
    MaPLe& 70.72& 93.53 & \textbf{90.49} & 65.57 & 72.23 & 86.20 & 24.74 & 67.01 & \textbf{46.49} & 48.06 & \textbf{68.69} & 66.30 \\
    KgCoOp & 70.66& 93.92 & 89.83 & 65.41 & 70.01 & 86.36 & 22.51 & 66.16 & 46.35 & 46.04 & 68.60 & 65.51 \\
     \midrule  
    PMPO& 70.78& 93.67 & 90.09 & \textbf{65.64} & \textbf{72.82} & \textbf{86.55} & \textbf{25.10} & \textbf{67.54} &  46.38 & \textbf{49.28} & 68.46 & \textbf{66.94} \\
    \midrule 
    \bottomrule
    \end{tabular}%
  \label{tab:cross_evalution}%
\end{table*}

\begin{figure*}[htbp]
\begin{center}
\vspace{-0.2cm}
\includegraphics[width=1.00\linewidth]{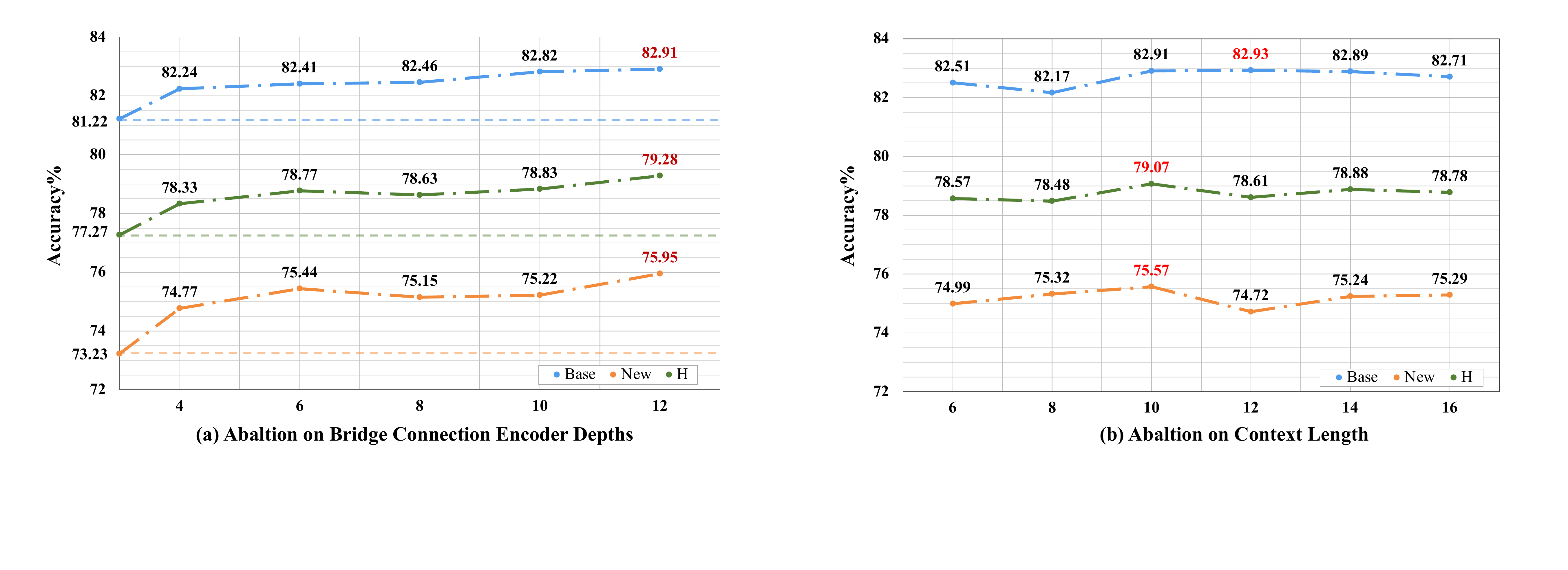}
\end{center}
\vspace{-1.5cm}
\caption{Ablation the bridge connection depths and contextual length, The results are the average of 11 datasets' accuracy }
\vspace{-0.3cm}
\label{fig:ablation_depth}
\end{figure*}

\textbf{Bridge Connection Depths and Context Lengths:}
In Figure.\ref{fig:ablation_depth}, we examine the impact of varying the depths, which bridge the text encoder prompts and visual encoder blocks. For this ablation study, the number of prompts ($N$) is set to 4. We obtain a unimodal multi-prompt model when the depth is set to $0$. This vanilla model directly matches the multi-prompt ensemble with the visual features, resulting in the lowest average performance across the 11 datasets, with a base accuracy at $73.23\%$, a new accuracy at $73.23\%$, and a harmonic mean ($H$) of $81.22\%$. In contrast, the default setting with depths set to 12 outperforms the vanilla setting by $2.72\%$ on the base classes, $2.01\%$ on the new classes, and $1.69\%$ on $H$. This improvement might be attributed to the vanilla model's tendency to learn trivial solutions without comprehensive prompts. This comparison demonstrates that our partitioned cross-modal design significantly enhances the performance of multi-prompt learning.

\textbf{Ensembling Manually Designed Prompts:}
As shown in Figure \ref{tab: ablation_m2}, we evaluate the PMPO in two different settings across 11 datasets: without ensembling the manually designed prompt and with the ensemble (default). In terms of results, the setting without manually designed prompts achieves better average accuracy on the base classes. However, the setting with manually designed prompts demonstrates stronger generalization performance on the new classes (better on 10 out of 11 datasets than the setting without manual prompts). Therefore, we choose the setting with manually designed prompts as our default configuration.
\begin{table*}[htbp]
  \caption{Effect of number of shots on the base-to-new  generalization on $11$ datasets. $M$ means  the manually-designed prompt. H:Harmonic mean.}
  \centering
  
    \setlength{\tabcolsep}{10pt}
    \begin{tabular}{lccc|ccc|ccc}
    \toprule
    \multicolumn{1}{l}{} & \multicolumn{3}{c|}{Base}  & \multicolumn{3}{c|}{New}&  \multicolumn{3}{c}{H}  \\
    \cmidrule{2-10}
    \multicolumn{1}{l}{datasets} & \multicolumn{1}{c}{4-shot}  & \multicolumn{1}{c}{8-shot}& 
    \multicolumn{1}{c|}{16-shot}& \multicolumn{1}{c}{4-shot}& \multicolumn{1}{c}{8-shot} & 
    \multicolumn{1}{c|}{16-shot} & \multicolumn{1}{c}{4-shot} & \multicolumn{1}{c}{8-shot} &
    \multicolumn{1}{c}{16-shot} 
    \\
     \midrule   
     CoOp & 78.43 & \textbf{80.73} & 82.69 &68.03 &68.39 &63.22 & 72.44 &73.50 &  71.66 \\
     CoCoOp & 76.72 & 78.56& 80.47& \textbf{73.34} &72.00& 71.69& 74.85 & 74.90  &  75.83\\
     MaPLe & 76.71 &79.12 &82.27 & 72.86& \textbf{74.74} & 75.14 &74.70& 76.93 &  78.55 \\
     KgCoOp & \textbf{79.92} &78.36 &80.73 &73.11 &73.89 & 73.60& \textbf{75.90} & 76.60 & 77.00 \\
     \midrule 
     PMPO & 78.34& 80.25&\textbf{82.91} & 73.05 & 74.11 & \textbf{75.95}&75.59 & \textbf{77.06} & \textbf{79.27} \\
    \midrule         
    \bottomrule
    \end{tabular}%
  \label{tab: ablation_shot}
\end{table*}


\textbf{Number of Shots:} We investigate the average performance achieved with various K-shot samples across 11 datasets under base class-to-new class settings, as shown in Figure \ref{tab: ablation_shot}. We sample K shots with K set to 4, 8, and 16 (default). As the number of shots increases, the PMPO consistently performs better, with the harmonic mean rising from ($75.59\%$ to $79.27\%$). The PMPO achieves the best results under the 8-shot and 16-shot settings, with superior scores of $76.93\%$ vs. $77.06\%$ and $78.55\%$ vs. $79.27\%$, respectively. For the 16-shot setting, the PMPO attains state-of-the-art performance for both the base and new classes, as well as the harmonic mean. However, the PMPO's performance is slightly lower in the 4-shot setting ($75.59\%$ vs. $75.90\%$). This can be attributed to the PMPO's requirement to learn more prompts, and the limited low-shot samples are insufficient for effectively learning discriminative prompts.

\textbf{Comparison with Alternative Designs:}
In Table \ref{tab: ablation_modal}, we present a simple comparison with other prompting designs. Since CoOp is defined as a uni-prompt, unimodal method, we introduce two alternative methods: PMPO-MS, which uses multi-prompts (including manually designed prompts) to replace the uni-learnable prompt and lacks a bridge connection between the text and image encoders, and PMPO-SM, which features a uni-prompt with visual depth connections (one learnable prompt has connections with 12 visual transformer blocks). The results show that both PMPO-MS and PMPO-SM underperform compared to the multi-prompt, partitioned, cross-modal PMPO approach.

\textbf{Model Complexity:}
In Table \ref{tab: ablation_parameters}, we compare the model complexities of different methods. We use the best performance setting, with 4 prompts, a length of 10, and a depth of 12. Because the PMPO is a multi-prompt framework, the training parameters of the PMPO are significantly more numerous than those of the other methods (4.72 M vs. 2048 for CoOp). Moreover, the inference speed is also slower than that of the baseline CoOp method. This situation is a typical problem faced by multi-prompt learning and will be explored in future work. Notably, the training effectiveness is significantly improved because only 6 epochs are needed compared to the 200 epochs required by CoOp.

\begin{table}[htbp]
  \caption{Comparison of model complexity of different prompt learning methos. KgCoOp is comparable to the CoOp.}
  \centering
    \setlength{\tabcolsep}{8pt}
    \begin{tabular}{lcccc}
    \toprule
    \multicolumn{1}{l}{Method} & \multicolumn{1}{c}{Params} & \multicolumn{1}{c}{Params (\%CLIP)}  & \multicolumn{1}{c}{FPS}&  \multicolumn{1}{c}{H}  \\
    \midrule 
     CoOp & 2048 & 0.002 &1353.0 & 71.66\\
     CoCoOp  & 35360 & 0.03 &  15.1 & 75.83  \\
     MaPLe  &3.55M & 2.85 & 1365.1 & 78.55  \\
     \midrule 
     \rowcolor{gray!20} PMPO & 4.72M & 3.79 & 1107.9 & \textbf{79.27} \\
    \midrule         
    \bottomrule
    \end{tabular}%
  \label{tab: ablation_parameters}
\end{table}
\label{section:ablation}
\subsection{Cross-Dataset Transfer Evaluation}
In Table \ref{tab:cross_evalution}, we assess the cross-dataset transfer generalization capability of the PMPO by training the model on all 1000 classes of ImageNet\cite{deng2009imagenet} and testing its performance on 10 other diverse datasets. The results indicate that the PMPO outperforms the other methods in 6 out of the 10 cross-dataset evaluation settings, with the highest average performance across all 11 datasets. The range of superior performance varies, from a $0.18\%$ improvement (on SUN397 compared with CoCoOp) to a $0.59\%$ enhancement (on Flowers102 compared with MaPLe). These findings underscore the competitive generalization ability of the PMPO approach, highlighting its adaptability across various datasets and settings.

\subsection{Domain Generalization}
Domain generalization aims to assess the generalization ability of a model by evaluating the trained model on a target dataset. Given CLIP's robust domain generalization performance, we also evaluate the PMPO's domain generalization ability by performing prompt tuning on the few-shot ImageNet dataset and testing the PMPO on ImageNetV2, ImageNet-Sketch, ImageNet-A, and ImageNet-R. These datasets share the same classes but have different data distributions than the ImageNet domain. As shown in Table \ref{tab: domain}, although CoOp achieves the highest performance on the ImageNet source dataset, its generalizability to a broader domain is weaker than that of CoCoOp. MaPLe and KgCoOp are also competitive methods, with MaPLe achieving the best performance among the competitors in 3 out of 4 domain generalization settings. However, the PMPO outperforms these methods on 3 target datasets. This comparison confirms that the partitioned multimodal prompts in the PMPO exhibit better domain generalization capabilities.

\section{Limitations}
Although the PMPO demonstrates competitive performance, it should be noted that partitioned multi-prompt learning has two limitations. 1) The PMPO demands more GPU memory than uni-prompt learning. That is because the PMPO needs multiple forward passes to the text encoder. This becomes significant as the number of prompts increases. This results in linear growth in the computational cost for the text encoder with respect to the number of classes. Future research may need to explore class sampling strategies to mitigate this issue. 2) The PMPO needs more shots for training than the other prompt tuning methods, as shown in Table \ref{tab: ablation_shot}. Some data argument methods may help mitigate this issue.
\section{Conclusion}
In this paper, we introduce the PMPO, a novel approach that addresses the limitations of uni-prompt learning in large-scale VLP models. By partitioning the depths of a visual encoder and connecting learnable prompts with the divided visual depths, our method enables different prompts to capture the hierarchical contextual depths of the observed visual representations.

Through extensive experiments conducted on 11 diverse image recognition datasets, we demonstrate that our proposed method, the PMPO, achieves outstanding performance across new class generalization, cross-dataset evaluation, and domain generalization tasks.

{\small
 \bibliographystyle{ieee_fullname}
 \bibliography{sample-base}
 }


\end{document}